\documentclass[a4paper, 10pt, conference]{ieeeconf}      

\IEEEoverridecommandlockouts                              

\usepackage{graphicx}
\usepackage{subcaption}
\usepackage{amsmath,amssymb,amsfonts}

\title{\LARGE \bf
Learning a Curve Guardian for Motorcycles
}

\author{Simon Hecker$^{*1}$, Alexander Liniger$^{*1}$, Henrik Maurenbrecher$^{1}$, Dengxin Dai$^{1}$, and Luc Van Gool$^{1,}$$^{2}$
\thanks{$^{*}$The authors contributed equally.}%
\thanks{$^{1}$ETH Zurich in Switzerland. $^{2}$KU Leuven in Belgium.}%
}

\begin{document}

\maketitle
\thispagestyle{empty}
\pagestyle{empty}

\begin{abstract}
Up to 17\% of all motorcycle accidents occur when the rider is maneuvering through a curve and the main cause of curve accidents can be attributed to inappropriate speed and wrong intra-lane position of the motorcycle. Existing curve warning systems lack crucial state estimation components and do not scale well. We propose a new type of road curvature warning system for motorcycles, combining the latest advances in computer vision, optimal control and mapping technologies to alleviate these shortcomings. Our contributes are fourfold: 1) we predict the motorcycle's intra-lane position using a convolutional neural network (CNN), 2) we predict the motorcycle roll angle using a CNN, 3) we use an upgraded controller model that incorporates road incline for a more realistic model and prediction, 4) we design a scale-able system by utilizing HERE Technologies map database to obtain the accurate road geometry of the future path. In addition, we present two datasets that are used for training and evaluating of our system respectively, both datasets will be made publicly available. We test our system on a diverse set of real world scenarios and present a detailed case-study. We show that our system is able to predict more accurate and safer curve trajectories, and consequently warn and improve the safety for motorcyclists. 

\end{abstract}

\section{INTRODUCTION}
The Department of Transportation National Highway Traffic Safety found that 37461 people were killed due to accidents on US roads in 2016. Of these, 5286 were motorcyclists, constituting roughly 14\,\% of all fatalities, even though motorcycles account for less than 1\,\% of the combined traveled distance of all road vehicles. The Hurt Report (1981) studied the cause of fatal motorcycle accidents and concluded that in approximately 25\,\% of accidents only the motorcycle was involved and of these roughly $2/3$ were caused by an error of the rider. The Maids Report \cite{maids2004depth}, published in 2004 came to a similar conclusion by analyzing 921 powered two-wheeler accidents, across five different European countries. Accidents were assessed based on 2'000 different variables, and the study found that in 37\% of these accidents, an error of the rider was the main cause and that in 30\% of the accidents, the rider did not have enough time to avoid the accident.

The most difficult maneuver on a motorcycle is riding through a curve, as it involves the rider adjusting their speed and the lane position of the vehicle before entering the turn and subsequently adjusting the handlebars to properly roll the motorcycle. The slightest mistake can lead to an accident and it is thus no surprise that studies have shown, that many motorcycle accidents (25-30\,\%) \cite{article}, occur when the rider is negotiating a curve. This type of accident has been found to be responsible for around 15\,\% \cite{article} of all rider fatalities.
Further investigations have revealed, that most curve related accidents are avoidable, as they are mainly the result of excess speed and inappropriate motorcycle lane position that can result in a slide-out, a fall as a consequence of over-braking, running off the lane or under-cornering \cite{biral2014experimental}.

These numbers stress the need for the development of an Advanced Rider Assistance Systems (ARAS) that can help a motorcyclist navigate safely through a curve. Such a system would require four main components to function: 1) accurate road geometry of the future path, 2) accurate localization of the motorcycle within the lane, 3) accurate motorcycle roll angle prediction and 4) a realistic controller, motorcycle and road model for trajectory planning.

Indeed, in 2010, \cite{biral2010intelligent, biral2014experimental} proposed the first curve warning system by calculating the optimal future maneuver based on the upcoming road geometry and the vehicle's current state. However, while their system performed well, they were unable to include all four components completely and relied on assumptions for several of their state estimations. First, they always assumed the motorcycle position to be in the center of the lane, which does not hold true in practice. The intra-lane position can have a significant effect on the safety of a subsequent curve trajectory, thus determining the accurate intra-lane position is vital for a fully functional curve warning system. Second, the roll angle was predicted using high precision IMUs adding to the overall cost of the system and posing a challenge in highly dynamic environments. Third, their controller did not include road incline and roll dependent lane limits which, especially for motorcycle maneuvers, can have a significant impact on the selection of safe trajectories. Finally, they manually measured the road geometry, preventing scaling the system to large road networks.

Subsequent works did try to alleviate some of these shortcomings such as using inverse perspective mapping to determine the road curvature on-the-fly from images \cite{damon2018image} or more reliably estimating motorcycle roll \cite{boniolo2009roll, lot2012real}, but these, again, employed expensive, high precision IMUs.

Using the recent advances in computer vision, the availability of large scale industry standard maps and an improved controller design, we propose a system that can overcome the shortcomings of the system designed by \cite{biral2010intelligent}. Our system: 1) uses a convolutional neural network (CNN) to predict the intra-lane position a novelty, 2) it employs a second CNN to estimate the current motorcycle roll angle removing the need for expensive IMUs, 3) it uses an upgraded controller model that incorporates road incline and roll dependent lane constraints for a more realistic predictions 4) it queries HERE Technologies map database to obtain the accurate road geometry of the future path allowing our system to scale to roads in the Americas, Asia and Europe.

\section{RELATED WORK}

\subsection{Computer Vision for Autonomous Driving}
The recent advances in CNNs and mapping technology have lead to an explosion in applications for the automotive sector. Cars have received most of this attention, with full CNN based driving models being developed, which utilize camera images and visual \cite{hecker2018end} or numeric \cite{driving:hecker2019learning} maps. Likewise CNNs have been used as guardian angle systems for cars such as to predict failure of the underlying driving system \cite{driving:failure:prediction}. For motorcycles specifically there has been less activity but nonetheless the field is benefiting from the technological improvements.
\cite{damon2018image} were the first to demonstrate image-based lateral position estimation within a lane, however they rely on traditional, geometric computer vision techniques. We believe formulating the task as a learning problem and utilizing the advances in CNNs will create a perception system that is able to generalize better for a diverse set of scenarios.
Image-based roll estimation for motorcycles has been proposed by \cite{damon2018inverse, schlipsing2011video}. However, again, these approaches used traditional, geometric computer vision techniques with all their shortcomings. Image rotation prediction with CNNs has been demonstrated before \cite{fischer2015image}, but never explicitly for automotive domain applications.

\subsection{Control Theory}
In this work we use an optimal control approach to predict the optimal future trajectory of a motorcycle. Based on this predicted trajectory we decide if the current driving situation should trigger a warning. Several papers propose such a system for both cars and motorcycles \cite{biral2010intelligent,biral2014experimental,Bosetti2015,schwarting2017}. However, fundamentally all these approach rely on model-based predictive controllers. Thus, prediction based warning systems also relate to model predictive controllers, which is a common control technique for cars \cite{falcone2007,Liniger2015,lot2014}, however, for motorcycles far fewer solutions exist but notable are \cite{frezza2004,Hauser2006,saccon2008,Chu2018}.

\section{APPROACH}
In this section we present our contributions to improve existing curve warning systems by 1) learning intra-lane localization, 2) learning roll angle estimation, 3) using industry standard maps for accurate future road geometry, and 4) reformulating the optimal control problem to include road incline and rider pose. We also highlight our two dataset contributions, namely our \textit{Learning Dataset}, recorded with a car and used to train our two CNNs and our \textit{Motorcycle Dataset} recorded with a motorcycle and used to evaluate our curve warning system. 

\subsection{Datasets}
\begin{figure}
    \centering
    \begin{subfigure}[b]{0.6\linewidth}
        \includegraphics[width=\textwidth]{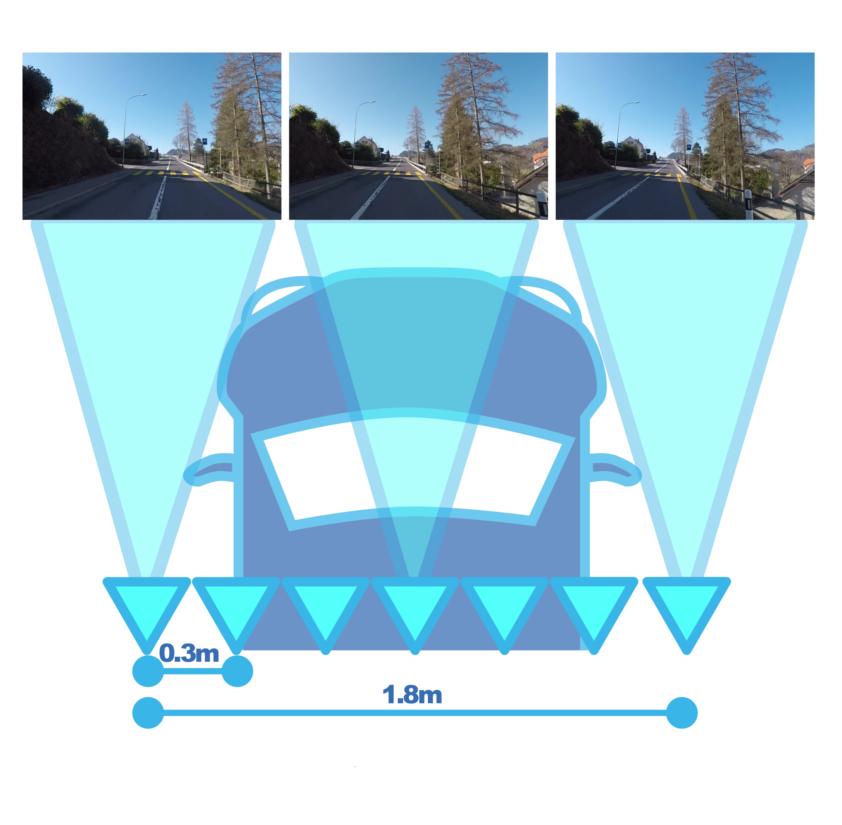}
        \caption{Car}
        \label{fig:car_rig}
    \end{subfigure}
    ~ 
    \begin{subfigure}[b]{0.22\linewidth}
        \includegraphics[width=\textwidth]{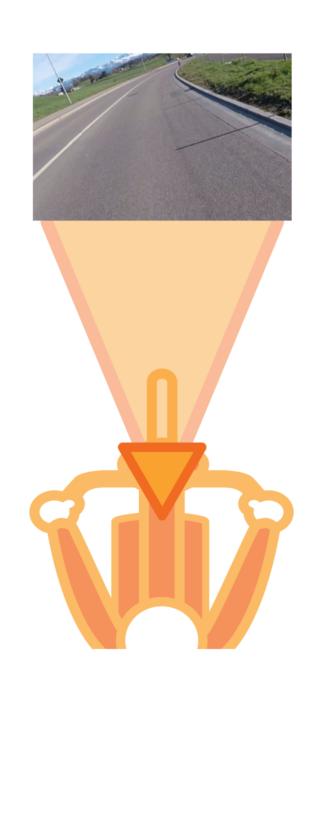}
        \caption{Motorcycle}
        \label{fig:motorcycle_rig}
    \end{subfigure}
    ~ 
    \caption{Camera rig configuration for (a) \textit{Learning Dataset} and (b) \textit{Motorcycle Dataset}. Note that in reality all 7 cameras of (a) record.}
    \label{fig:rig_configuration}
\end{figure}

We first introduce our two datasets and how we include industry standard HD maps from HERE Technologies to enrich our \textit{Motorcycle Dataset}.
\subsubsection{Learning Dataset}
Our networks are trained by either supplying images recorded at different locations within a lane or, in addition, augmented via a rotation to simulate and learn to predict motorcycle roll. It would be difficult to collect large-scale accurate lane location data using a motorcycle as this would require hours of riding at a particular location within the lane. To solve this problem, we design and build a camera rig, shown in Fig. \ref{fig:car_rig}, that allows us to simultaneously capture images from different lane positions. Using this setup we can accurately capture images spaced at defined intervals within the lane. The camera rig boasts seven front-facing GoPro Hero 5 Black cameras that are equally spaced 30cm apart and concurrently record the lane in front. The GoPro cameras deliver images at 60Hz with 1080p resolution. The camera rig is mounted on the roof of a car, and with this configuration a total of 700km (15hrs per camera) are recorded by driving on various road types in Switzerland. To improve the localization ability of the network, the driver is tasked with maneuvering the vehicle as close to the lane-center as possible and avoid overtaking. 

The \textit{Learning Dataset} is used solely to train our two neural networks (intra-lane position prediction and roll prediction) and it will be made available to the community. It contains a total of around 2 million images from seven different lane positions (we extract images at 10Hz from the video), and is split into a training, validation and test set with 1.5m, 100k and 400k images respectively. The seven camera positions $n$ can be used to train a lane localization network, whereas augmenting any image within the dataset by a rotation $\phi$ allows the training of a roll prediction network.

\subsubsection{Motorcycle Dataset}
To allow for a more realistic evaluation of the complete system, in particular the controller, we proceed with collecting a second dataset named \textit{Motorcycle Dataset}, the rig configuration is shown in Fig. \ref{fig:motorcycle_rig}. It contains 4 hours of real world motorcycle riding recorded by a front facing GoPro Hero 5. Images are extracted exactly as in the \textit{Learning Dataset}, however, we also record our GPS coordinates during acquisition.  We use a path-matcher, based on a hidden markov model employing the Viterbi algorithm \cite{forney1973viterbi}, to calculate the most likely path traveled by the motorcycle during recording, snapping the GPS trace to the underlying road network. This improves our localization accuracy significantly, especially in urban environments where the GPS signal may be weak and noisy. Through the path-matcher we obtain a map matched GPS coordinate for each synchronized video frame, which is then used to query the HERE Technologies map database to obtain the road curvature, incline and speed limit at that position. We fuse this collected map data into our \textit{Motorcycle Dataset} and can thus obtain accurate road attributes required as state space variables by the controller for the subsequent curve warning evaluation.

\subsection{Learned Localization}
A key component for a well performing curve warning system for motorcycles is the intra-lane localization - how far the motorcyclist is to the left or right of the lane center. This position has a large influence on the future trajectory and is significant to perform a safe curve negotiation. Not only is it important in terms of the motorcycle sliding out, but also to prevent the riders head from crossing onto the other lane, and potentially oncoming traffic, given a large roll of the motorcycle. 

All previous curve warning system approaches, such as \cite{biral2010intelligent}, have assumed a lane-center position for the motorcycle and this is correctly identified as a major shortcoming of the devised system. Recent advances in computer vision and deep learning now allow for accurate intra-lane localization from images. Consequently we present our intra-lane localization network called \textit{LNet}.

\subsubsection{Network Architecture}
We formulate our learning objective as a regression task on the seven discretized camera locations $n$ of the camera rig. Given an input image $I_n$ from one of the seven cameras, our network is tasked to predict the corresponding camera position $n$. By formulating our learning task as a regression, our network is able to interpolate lane positions between and beyond the discretized camera locations of the rig, and can thus generalize over all intra-lane positions encountered by a motorcycle. We have also experimented with formulating the problem as a classification, however this did not lead to satisfactory results and makes interpolating between the discretized positions more difficult. We base our network, \textit{LNet}, on the Resnet34 architecture of \cite{resnet2016} and replace the final layer of the network by a single output neuron for the regression task.

\subsubsection{Training Procedure}
Appropriate pre-training image-augmentation is vital for the robustness of the network in subsequent more realistic motorcycle test scenarios of our \textit{Motorcycle Dataset}. Given that the images of our \textit{Learning Dataset} are captured from accurately car-mounted, parallel, front facing cameras, the network is very sensitive to camera yaw or pitch changes that occur more frequently in the \textit{Motorcycle Dataset}. Camera yaw and pitch can change quite rapidly and significantly when mounted on a motorcycle vs. a car due to lateral lane maneuvers and motorcycle shock compression/uncompression during braking and accelerating respectively. We thus employ a strategy of random perspective warping the training image to simulate a yaw or pitch change of up to 10$^{\circ}$, this significantly improves the performance of our network.

\subsection{Learned Roll}
Estimating the roll angle of a motorcycle in a dynamic environment is notoriously difficult and requires expensive IMUs. Nonetheless, being able to accurately estimate the current roll angle of the motorcycle is crucial for a curve warning system. As our system is already camera based due to the lane localization network, we thus propose utilizing a second CNN capable of predicting the roll angle given an input image. Henceforth, we will refer to this CNN as \textit{RNet}. As it is difficult to obtain real, ground-truth annotated data for this task, we choose to artificially rotate images of our \textit{Learning Dataset} to simulate data of a real motorcycle roll. Thus, our image rotation angle becomes our ground-truth motorcycle roll we seek to learn.

\subsubsection{Network Architecture}
Our network should be able to detect all possible roll angles of a motorcycle, consequently we define the roll angle range from $-90^{\circ}$ to $+90^{\circ}$. We agree that a realistic roll range is between +/- 60$^{\circ}$, however this full range makes our network more robust. We formulate our learning problem as a regression and optimize the mean absolute error between predicted image rotation and actual image rotation. In particular, given an image $I$ from our \textit{Learning Dataset}, belonging to any of the seven cameras, we artificially rotate the image by a random angle $\phi$ defined by our training procedure. The network is then presented $I_\phi$ and asked to predict $\phi$. We benefit from the superior interpolation properties when formulating our learning problem as a regression as opposed to a classification. The architecture of \textit{RNet} is identical to \textit{LNet}.

\subsubsection{Training Procedure}
Just as for lane localization training, careful image augmentation is required for successful training. In particular, all images must be rotated several times to prevent the network from simply learning the rotation induced interpolation artifacts from the target artificial rotation $\phi$, which would prevent real world use. Also the images must be center-cropped to 760 pixels to prevent black space or padding from becoming visible to the network due to the applied rotation. The pre-processing steps for the $i^{th}$ image $I_i$ are as follows:
\begin{enumerate}
    \item rotate $I_i$ by $\theta \in$[0$^{\circ}, 360^{\circ}$]
    \item rotate $I_i$ by $\phi_{i} \in$[$-90^{\circ}, +90^{\circ}$]
    \item rotate $I_i$ by $-\theta$
    \item center-crop $I_i$ to 760 pixel square
\end{enumerate}
Steps 1 and 3, a clockwise and subsequent counter-clockwise rotation by a random angle $\theta$ is necessary to prevent the network from simply learning the rotation induced interpolation artifacts of $\phi$.

\subsection{Optimal Control}
This section discusses the formulation of the optimal control problem and its implementation. The formulation is similar to \cite{biral2010intelligent}. However, we propose improvements of all the components and use a standard approach to convert the optimal control problem into a Nonlinear Optimization Problem (NLP). We will first introduce the model, including the influence of the slope, then formulate the constraints, where we introduce our roll dependent lane constraints. Finally, we describe the terminal constraints and cost, and combine them in our final optimal control problem. Note that we also improve the terminal constraints by including a terminal yaw rate, and use the maneuver time as a cost, which should remove unwanted artifacts that are caused by maximizing the velocity as proposed in \cite{biral2010intelligent}. Compared to \cite{biral2010intelligent} we use a standard multiple shooting approach to cast our optimal control problem into an NLP, which allows using standard modeling tools and mature optimization solvers, and would even allow using tailored real-time NMPC solvers such as \cite{zanelli2017}.

\subsubsection{Model}
In this paper we use the model proposed by \cite{biral2010intelligent}, however, additionally include the road slope due to the importance of the slope when driving in hilly and mountainous regions. The simplified motorcycle model assumes that the longitudinal and lateral motion are decoupled and that the motorcycle can be modeled as an free rolling disc, which cannot move laterally, but is allowed to yaw and roll. This model captures the most important dynamics of the motorcycle, while avoiding complications such as tire models, rider position and rider inputs. 

To limit the model to the region where it is valid, we further impose a combined acceleration constraint (discussed in the next section), similar to \cite{biral2010intelligent}. This acceleration constraint is essential since the non-slip assumption in the model is not realistic. But more importantly, the constraint allows us to limit the planned trajectory to motions the rider would be able to execute. 

 \begin{figure}[h]
 	\centering
 	\includegraphics[width=0.55\linewidth]{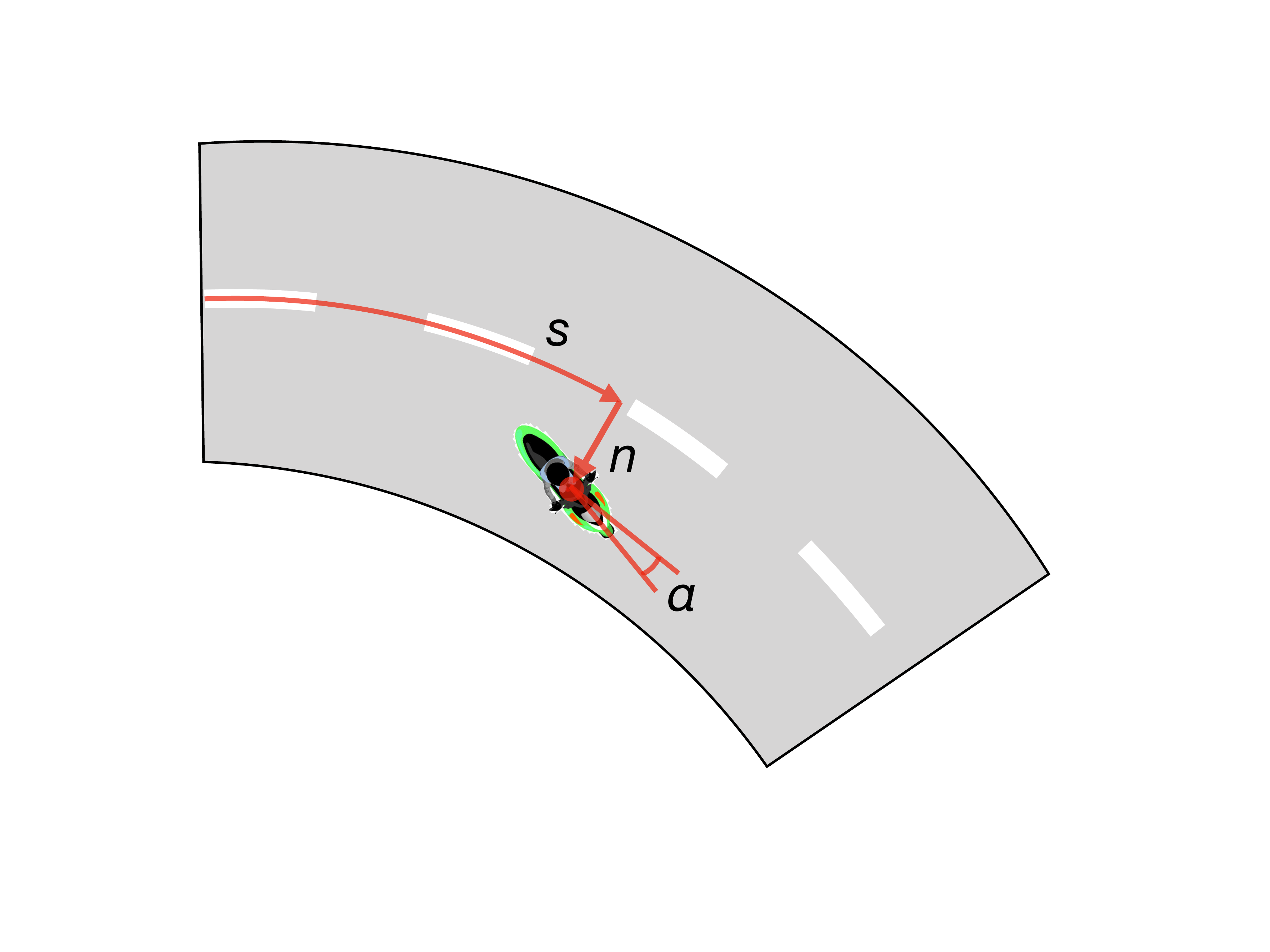}
 	\caption{Curvilinear coordinate system}
 	\label{fig:curvilinear}
 \end{figure}

We formulate our dynamics in a curvilinear coordinate system, see Fig. \ref{fig:curvilinear}, where the position and heading is transformed in a local coordinate system with respect to a known path, in our case the road lane divider. Furthermore, we assume that only the longitudinal jerk and the yaw jerk (rate of change of the yaw rate) can be controlled. Thus, the states of our system are $\hat{x} = [s,n,\alpha,\varphi,u_x,w_\psi,w_\varphi,a_x,a_\psi]^T$, where $s$ is the arc length along the road lane divider, $n$ the lateral distance to the lane divider, $\alpha$ the local heading, and $\varphi$ the roll of the bike. Furthermore, is $u_x$ the longitudinal velocity, $w_\psi$ the yaw rate, $w_\varphi$ the roll rate, $a_x$ the longitudinal acceleration and $a_\psi$ the yaw acceleration. The inputs are $u = [j_x,j_\psi]^T$ the longitudinal and the yaw jerk.

Additionally, we include the road slope by assuming small slope angles (similar to \cite{lot2014}) and only consider the road slope in the longitudinal dynamics. Under these assumption, and the following notation $\text{s}_\varphi = \sin(\varphi)$ and $\text{c}_\varphi = \cos(\varphi)$ used for readability, the dynamics of the motorcycle can be written as 
\begin{align} \label{eq:dynamics}
    \dot{s} &= \frac{u_{x} \cos \alpha }{1 - n \kappa(s)} \nonumber \\
    \dot{n} &= u_{x} \sin \alpha \nonumber \\
	\dot{\alpha} &=  w_{\psi} - \kappa(s) \dot{s} \nonumber \\
	\dot{\varphi} &= w_\varphi \nonumber \\
	\dot{u}_x &= a_x + g \sigma(s) cos(\alpha) \nonumber \\
	\dot{w}_\psi &= a_\psi \nonumber \\
	\dot{w}_\varphi &= h\frac{ ( g \text{s}_\varphi - w_{\psi} u_{x} \text{c}_\varphi + w_{\psi}^2 h \text{s}_\varphi \cos (\varphi))}{\rho_{x}^2 + h^2 + rh \text{c}_\varphi} \nonumber \\ & 
	\quad +  \frac{I_{w}}{m} \frac{w_{\psi} \text{c}_\varphi (w_{\psi} \text{s}_\varphi - u_{x}/R)}{\rho_{x}^2 + h^2 + rh \text{c}_\varphi} \nonumber \\ & 
	\quad +  r \frac{(h (w_{\varphi}^2 + w_{\psi}^2) \text{s}_\varphi - w_{\psi} u_x)}{\rho_{x}^2 + h^2 + rh \text{c}_\varphi} \nonumber \\
	\dot{a}_x &= j_x \nonumber \\
	\dot{a}_\psi &= j_\psi \,,
\end{align}

where, $\kappa(s)$ is the curvature of the road at $s$, $\sigma(s)$ the slope of the road at $s$. Furthermore, $g$ is the gravity constant, $h$ the height of the center of gravity, $r$ the tire cross section, $\rho_x$ the roll inertia radius, $R$ the radius of the wheel, $m$ the mass of the motorcycle including the rider, and finally $I_w$ the inertia of the wheel. From now on, we will refer to the dynamics in \eqref{eq:dynamics} as $\dot{\hat{x}} = \hat{f}(\hat{x},u)$.

To formulate the optimal control problem for our warning system we first transform the system into the space-domain, and then discretize the dynamics. Transforming the system into space is a popular approach to formulate optimal control problems in automotive applications \cite{lot2014}. The formulation is especially promising for long predictions and is often less dependent on the initial guess as time-domain formulations. Our time-domain dynamics $\dot{\hat{x}}(t) = \hat{f}(\hat{x}(t),u(t))$ can be transformed into the space-domain using the following transformation $\hat{x}'(s) = (1/\dot{s}) \hat{f}(\hat{x}(s),u(s))$. Note that this transformation makes the first state redundant allowing us to eliminate $s$. We denote the new space-domain system, with the removed $s$ state as $x'(s) =f(x(s),u(s))$, where $x$ is defined as $\hat{x}$ without $s$. 

The system can be discretized using an ODE discretization scheme, we choose an Euler forward approach resulting in the following discrete-space system, $x_{k+1} = x_k + d_s f(x_k,u_k)$, where $d_s$ is the discretization distance. 

\subsubsection{Constraints}
To limit the prediction of our warning system to maneuvers that are executable by the rider, and where the model assumptions are valid, we impose combined acceleration constraints, in vehicle dynamics these are often denoted as g-g constraints. The idea is to limit the longitudinal and lateral acceleration within an ellipse of appropriate shape and size. Therefore, note that the longitudinal acceleration is given by $a_{\text{long}} = a_x + g \sigma(s) cos(\alpha)$, as visible in \eqref{eq:dynamics}. The longitudinal acceleration is given by $a_{\text{long}} = u_x w_\psi$, by the no-slip assumption. Thus the combined acceleration constraint can be formulated as,
\begin{align}
    \left(\frac{a_{x,k} + g \sigma(s) \cos(\alpha_k)}{a_{x,\max}}\right)^2 + \left(\frac{u_{x,k} w_{\psi,k}}{a_{y,\max}}\right)^2 \leq 1\,, \label{eq:accCon}
\end{align}
where $a_{x,\max}$ and $a_{y,\max}$ are the maximum longitudinal and later acceleration, determined for the given rider. Note that the two parameters determine the shape of the ellipse and that in our study we use $a_{x,\max} = 4$\,m/s$^2$ and $a_{y,\max} = 7$\,m/s$^2$.

Additionally, it is necessary to constrain the motorcycle and rider to stay within the lane boundaries, also considering the roll angle of the bike, which is not considered in \cite{biral2010intelligent} but is crucial for the safety of the rider. This can be achieved using a constraint on $n$, the lateral deviation from the road lane divider, as well as the roll angle $\varphi$. By assuming small roll angles $\varphi$ the constraint can be conservatively approximated as the following constraint, 
\begin{align}
    \max(0,-\varphi_k h_r) \leq n_k \leq \min(b_\text{r}(s),b_\text{r}(s)-\varphi_k h_r)\,, \label{eq:roadCon}
\end{align}
where $h_r$ is the height of the rider and $b_\text{r}(s)$ is the width of the road at $s$.

Additionally, we consider the speed limits by imposing a constraint of the form $u_{x,k} \leq u_\text{limit}(s)$, as one of the bounds, which is imposed on all states and inputs. These bounds are summarize as $x \in \mathcal{X}(s)$ and $u \in \mathcal{U}$.

\subsubsection{Terminal Constraints}
At the end of the horizon the motorcycle should be in a state which allows to keep on driving. This is especially important since we do not consider the road geometry beyond the horizon. We follow the approach proposed in \cite{biral2010intelligent} and use the following terminal constraints, $n_T = b_\text{r}/2$, $\alpha_T = 0$, $w_{\phi,T} = 0$, $a_{x,T} = 0$, and $a_{\psi,T} = 0$, where the subscript $T$ indicates the terminal step in the prediction horizon. In addition to these constraints we also impose a terminal constraint on the yaw rate $w_\psi$, based on the curvature. The condition is derived as a steady state condition for the local heading state $\alpha$, $0 = w_\psi - \kappa(s) \dot{s}$, if we use the previous terminal conditions this equation simplifies to $w_{\psi,T} = \kappa(s) u_{x,T}/(1-b_\text{r}/2 \kappa(s))$. All the terminal constraints, including our newly proposed terminal yaw rate constraint are summarize as $\mathcal{X}_T$.

\subsubsection{Cost}
The cost function for our bike warning system consists of three parts. We minimize the maneuver time, but this minimum time objective is traded off with a cost on the combined acceleration, as used in the acceleration constraint \eqref{eq:accCon}, and a quadratic cost on the longitudinal and yaw jerk. Note that the two latter costs can be interpreted as comfort costs. Thus, to formulate the stage cost note that the maneuver time is given by $T = \sum_{k=0}^N d_s /\dot{s}_k$. Which allows us to formulate the three cost terms as,
\begin{align}
    J_t &= \sum_{k=0}^N q_t d_s \frac{1-n_k \kappa(s)}{u_k \cos(\alpha_k)} \nonumber\\
    J_a &= \sum_{k=0}^{N+1} q_a \left(\hspace{-0.1cm}\left(\frac{a_{x,k} + g \sigma(s) \cos(\alpha_k)}{a_{x,\max}}\right)^2 + \left(\frac{u_{x,k} w_{\psi,k}}{a_{y,\max}}\right)^2\right) \nonumber\\
    J_j &=  \sum_{k=0}^N r_{x} j_{x,k}^2 + r_\psi j_{\psi,k}^2\,.
\end{align}
Where, $q_t \geq 0$, $q_a \geq 0$, $r_x \geq 0$, and $r_\psi \geq 0$ are tuning weights that allow us to change the behavior of the predicted trajectory. 

\subsubsection{Optimization Problem}
Combining all the previous sections, we can formulate the optimization problem which is solved at each step,
\begin{align}
    \min_{\mathbf{x},\mathbf{u}}\; \;& J_t + J_a + J_j  \nonumber\\ 
    \text{s.t.}\;\;\; & x_0 = x(0) \nonumber\\
    &x_{k+1} = x_k + d_s f(x_k,u_k) \nonumber\\
    &\left(\frac{a_{x,k} + g \sigma(s) \cos(\alpha_k)}{a_{x,\max}}\right)^2 + \left(\frac{u_{x,k} w_{\psi,k}}{a_{y,\max}}\right)^2 \leq 1  \nonumber\\
    &\max(0,-\varphi_k h_r) \leq n_k \leq \min(b_\text{r}(s),b_\text{r}(s)-\varphi_k h_r)  \nonumber\\
    &x_k \in \mathcal{X}(s), \quad u_{k} \in \mathcal{U}  \nonumber\\
    &x_{N+1} \in \mathcal{X}_T, \quad k = 0,...,N\,. \label{eq:mpc}
\end{align}
Where, $x(0)$, is the state feedback estimated using \textit{RNet} and \textit{LNet}, as well as the GPS information. Note that this involves roll-correcting the lane prediction by \textit{LNet} due to the camera being mounted on the windshield, thus $n(0) = n_\textit{LNet} - h_c \sin(\varphi_\textit{RNet})$, where $h_c$ is the height of the camera mount. Finally, the curvature, slope and speed limits are queried from the HERE database, and $\mathbf{x} = [x_0,...,x_{N+1}]$ and $\mathbf{u} = [u_0,...,u_{N}]$ are the state and input trajectory.

The resulting NLP \eqref{eq:mpc} is formulated in Julia \cite{bezanson2017julia}, using the JuMP modeling framework. The optimal solutions $\mathbf{x}^*(s)$ and $\mathbf{u}^*(s)$ is then found using the NLP solver IPOPT \cite{waechter2009introduction}.

\subsection{Curve Warning System}
In order to quantify the hazard the rider will face in the upcoming section of road, a threat estimation method was implemented based on \cite{biral2014experimental}. To characterize the level of risk the rider is in, they propose to use the optimal longitudinal jerk $j_x^*$. The rider is mainly at risk if the velocity is too high for the road ahead, if this is the case a rapid deceleration is necessary to keep the motorcycle safe, which can be seen in a high negative jerk $j_x$. Thus, the risk is classified into three levels, associated with the upcoming maneuver:

\begin{enumerate}
	\item If $\vartheta_{1} \leq j_{x}^*(s)$, the upcoming maneuver is deemed to be safe.
	\item If $ \vartheta_{1} < j_x^*(s) < \vartheta_{2}$, this signifies an intermediate level of risk, where the rider does not need to take immediate action, but the maneuver is close to the limit of the rider's capability.
	\item If $j_{x}^*(s) \leq \vartheta_{2} $, the rider must take immediate action, as the upcoming maneuver is beyond the rider's ability and near the limit of tire adherence.
\end{enumerate}

The values for $\vartheta_{1}$ and $\vartheta_{2}$ were set at -0.1\,m/s$^3$ and -0.5\,m/s$^3$, respectively.  These thresholds were determined experimentally in \cite{biral2014experimental}, note that future work should examine how the risk assessment can be improved and automated.

\section{EXPERIMENTS}

We structure our evaluation of the system into three parts. 

First we evaluate both \textit{LNet} and \textit{RNet} on the test-set of our \textit{Learning Dataset}. This test-set consists of images that both networks have not encountered during the training procedure. In a second step, both networks are evaluated on real data by recording the rotation and translation of a GoPro to obtain real-world measured ground-truth. In a final step, we present a case study of the combined system for a road section in our \textit{Motorcycle Dataset}.

\subsection{Network Evaluations with Simulated Data}
\subsubsection{LNet} 
We train \textit{LNet} using the 1.5m images from the training set of our \textit{Learning Dataset} and train for 2 epochs on an NVIDIA Titan X GPU for around 24 hours. We use PyTorch as our deep learning framework, optimize with AdamOptimzer and set an initial learning rate of $10^{-5}$. After two epochs of training, \textit{LNet} has an absolute mean error of 0.358 on the \textit{Learning Dataset} test-set, translating to a 10.7\,cm absolute mean error in metric coordinates of the car camera rig, see Fig. \ref{fig:car_rig}. However this performance should be considered as an ideal scenario, keeping in mind that all images of the test set are captured from the roof of a car with significantly less vibration and perspective warping (yaw/pitch) as found on a motorcycle. 

\subsubsection{RNet}
We train \textit{RNet} with the same training procedure as \textit{LNet}. After two epochs of training, \textit{RNet} achieves an absolute mean error of $1.5^{\circ}$ when predicting the artificial rotation of an image from the \textit{Learning Dataset} test-set.

Both networks perform very well on the test-set. But, this is to be expected and does not necessarily translate to good real-world performance when exposed to the domain shift of real-motorcycle data. Consequently, we must evaluate both networks on more realistic data, while still being able to obtain the ground-truth. 

\subsection{Network Evaluations with Real Data}
It is very difficult to obtain ground-truth lane positions of a moving motorcycle. This would require careful riding on preset marked lane positions and heavy annotation. Likewise, it is quite difficult to obtain ground-truth roll angle measurements from an IMU during motorcycle operation. We nonetheless want to evaluate our system on realistic data therefore devising the following strategies for both networks. 

\subsubsection{LNet} 
\begin{figure}[ht!]
 	\centering
 	\includegraphics[width=0.95\linewidth]{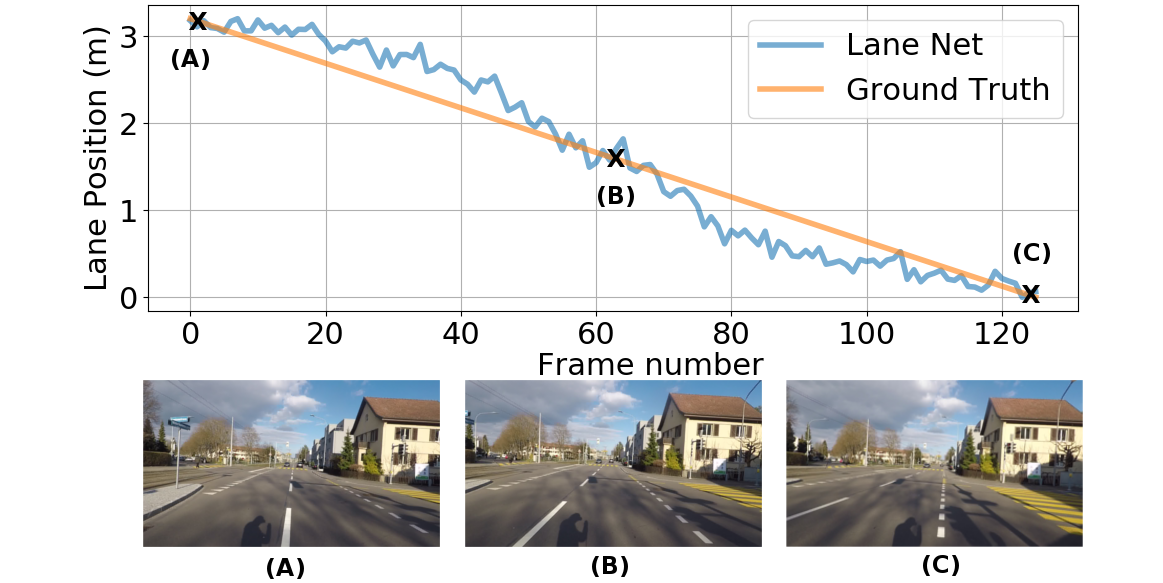}
 	\caption{Predictions of \textit{LNet} compared to the measured ground truth in meters. Sample images shown for locations (A), (B) and (C).}
 	\label{LaneNet_streettest}
 \end{figure}

A more realistic data test was conducted by using a hand-held GoPro to capture images from a lane in Zurich while walking across a pedestrian crossing. Using the stripes and gaps as guidance, a video was taken traversing the lane at constant velocity. This resulted in 126 frames for the entire lane crossing with frames being evenly spaced at around 2.5\,cm along the entire 3.2\,m lane width. This can be seen in Fig. \ref{LaneNet_streettest}, which shows the prediction of \textit{LNet} compared to the ground truth of the recording assuming constant velocity. Using this approach we can calculate the absolute mean error of our predictions to the ground truth which turns out to be 22.4\,cm.

\subsubsection{RNet}
\begin{figure}[ht]
	\centering
	\includegraphics[width=0.95\linewidth]{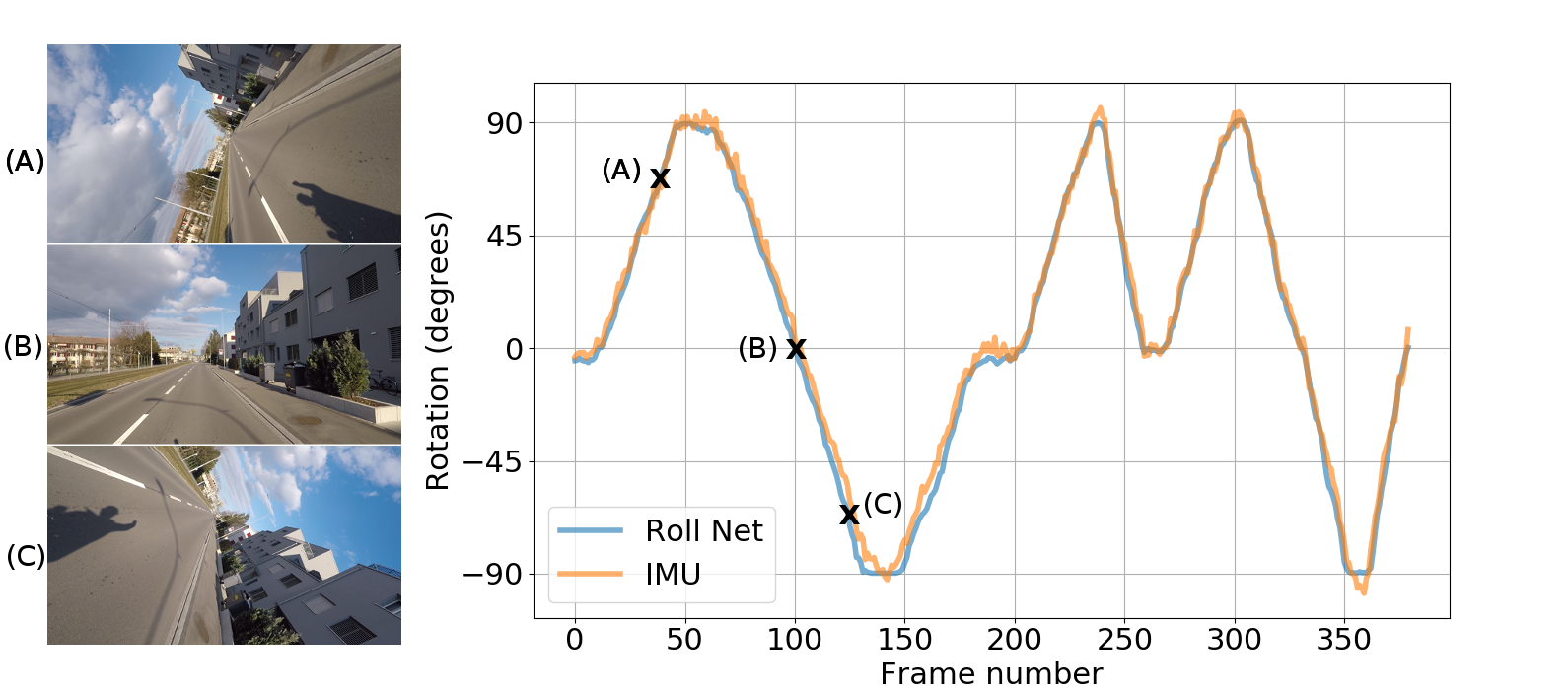}
	\caption{Predictions of \textit{RNet} compared to the IMU ground truth in degrees. Sample images shown for locations (A), (B) and (C).}
	\label{Streettest_RotNet}
\end{figure}

A realistic data test was conducted using a hand-held GoPro and manually rotating the camera while filming a street in Zurich. Using the GoPro's built-in IMU, the true roll angle of each taken frame $\varphi_{g}$ can be calculated. Note, that the camera was set to be level in terms of pitch. Due to the absence of vibrations from the Motorcycle, and the only motion stemming from the camera rotation, the IMU measurements are quite accurate and can be used to calculate the roll ground-truth. Using this method, our network can predict angles with a $3.7^{\circ}$  mean absolute error compared to the IMU measurements. An excerpt of the images used and the results of the test for \textit{RNet} can be found in Fig. \ref{Streettest_RotNet}. \textit{RNet} performs very well as shown by the almost identical roll predictions.

\subsection{Case Study}
\begin{figure*}[]
	\centering
	    \begin{minipage}{.45\textwidth} 
	        \centering
        	\includegraphics[width=.9\linewidth]{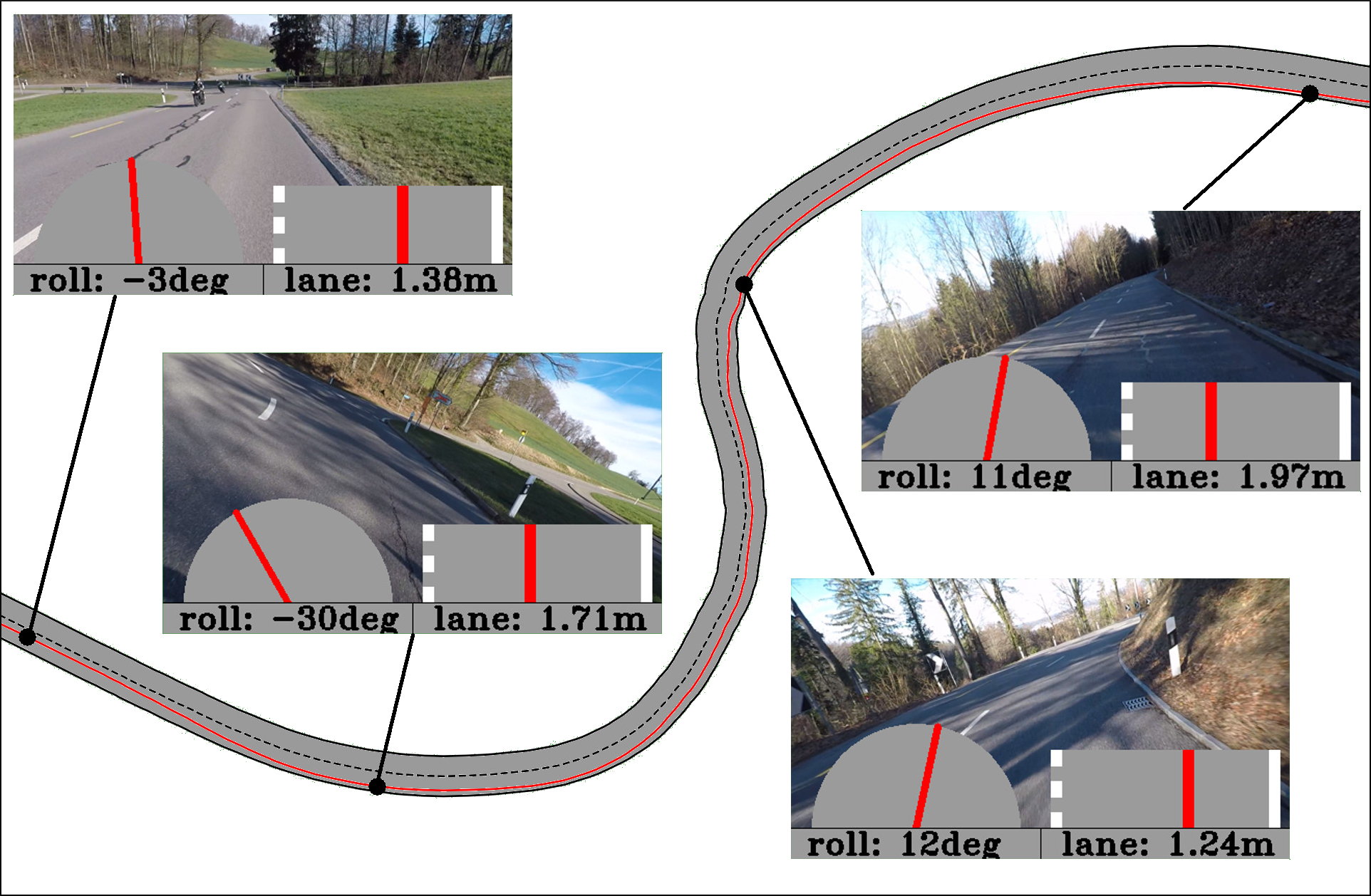}
        	\caption{The section of road used in our case study. \textit{Rnet} and \textit{LNet} predictions for four sample frames are shown, along with the optimal trajectory planned by the controller in red.}
        	\label{fig:ablation_traj}
    	\end{minipage}\hspace{1.5cm}%
    	\begin{minipage}{.45\textwidth}
	        \includegraphics[width=0.99\linewidth]{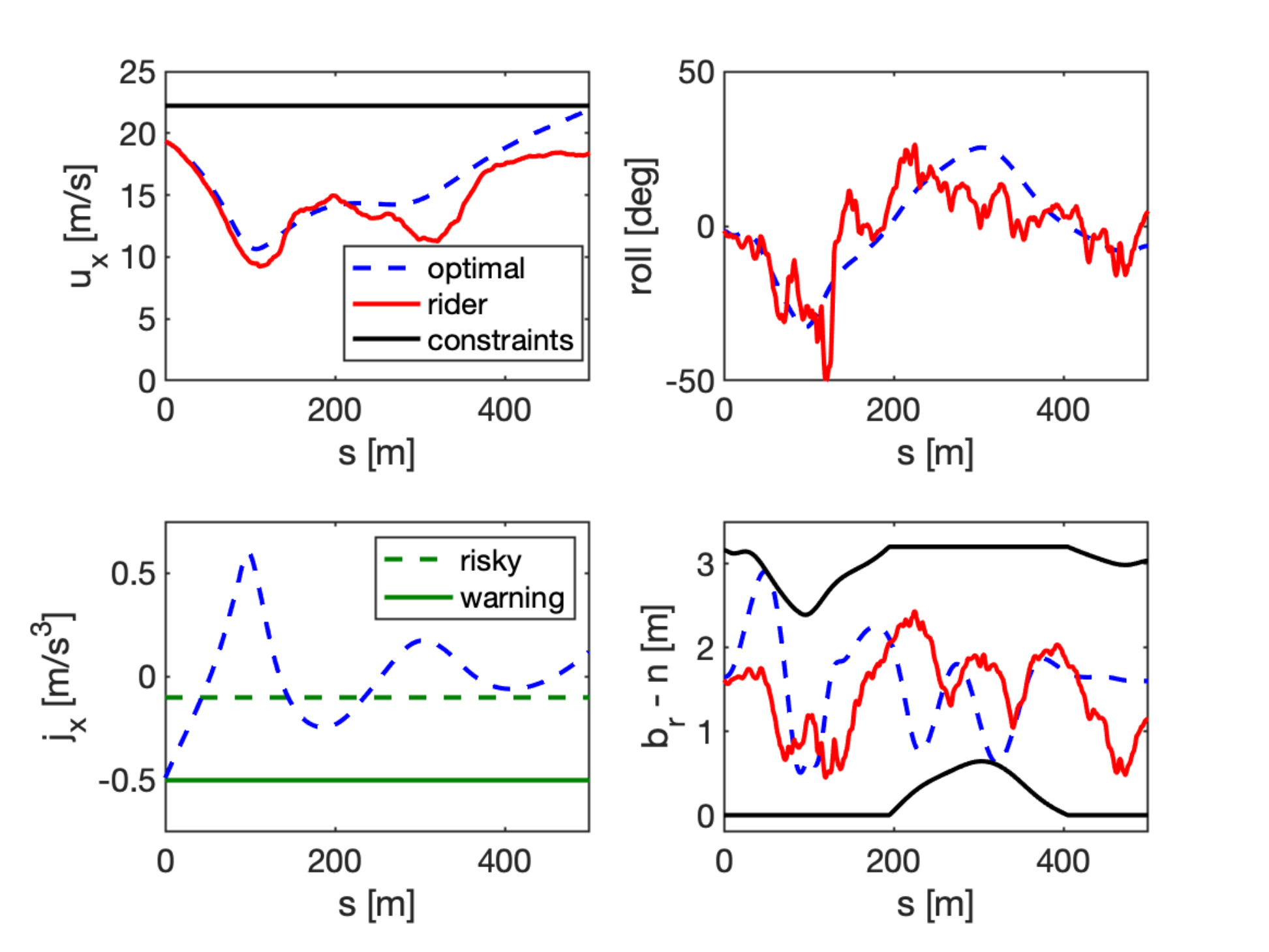}
	        \caption{Curve warning analysis, comparing optimal solution for $u_x$, $\varphi$, and $n$ with the motion of the rider, as well as the jerk $j_x$ and the warning thresholds.}
	        \label{fig:trajPlot}
    	\end{minipage}
    	\begin{minipage}{.45\textwidth}
	        \includegraphics[width=0.95\linewidth]{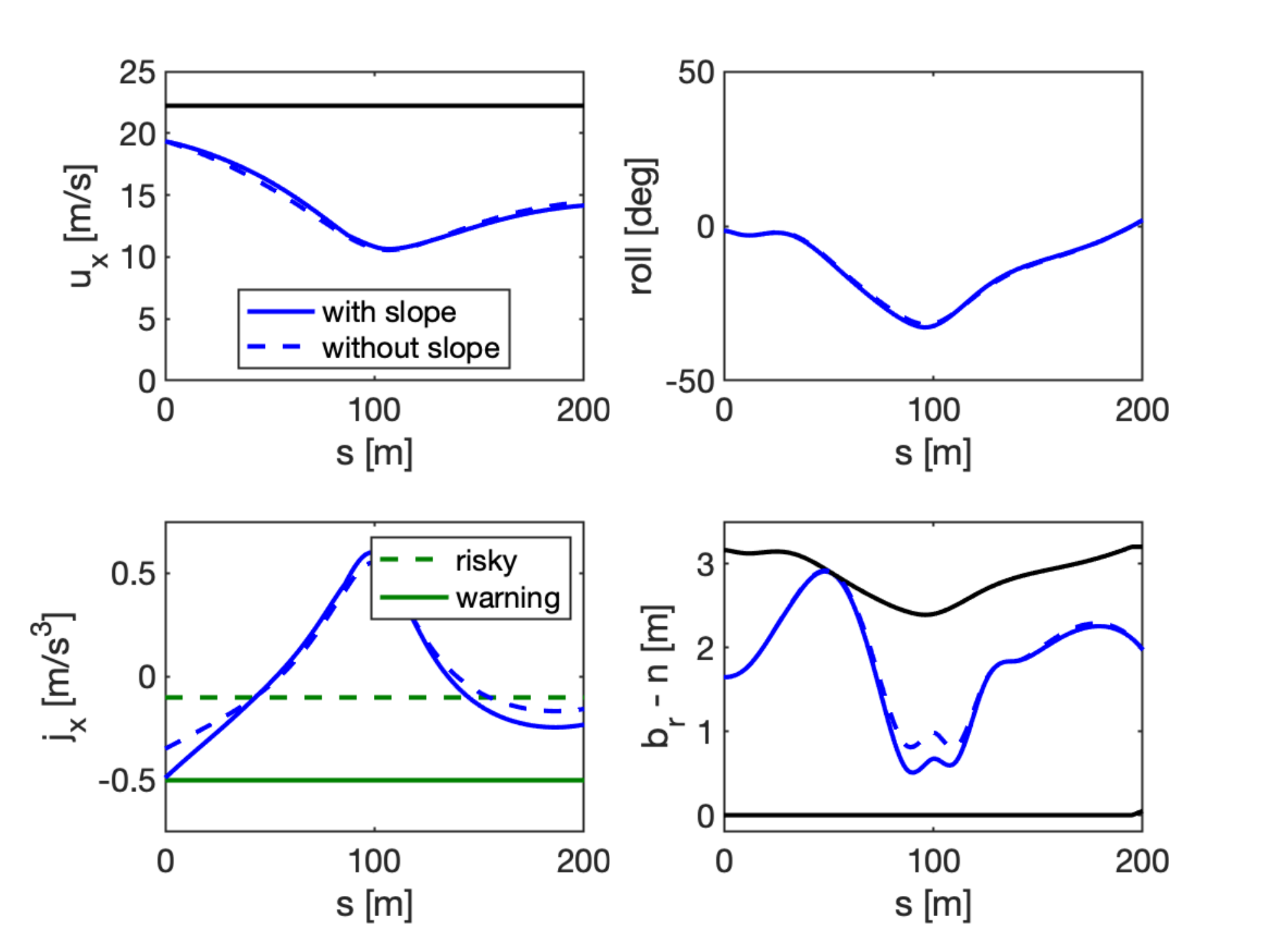}
        	\caption{Analyzing the influence of the road slope on the optimal solution. Neglecting the slope reduces the estimated risk.}
	        \label{fig:trajPlot_slope}
    	\end{minipage}\hspace{1.5cm}%
    	\begin{minipage}{.45\textwidth}
	        \includegraphics[width=0.95\linewidth]{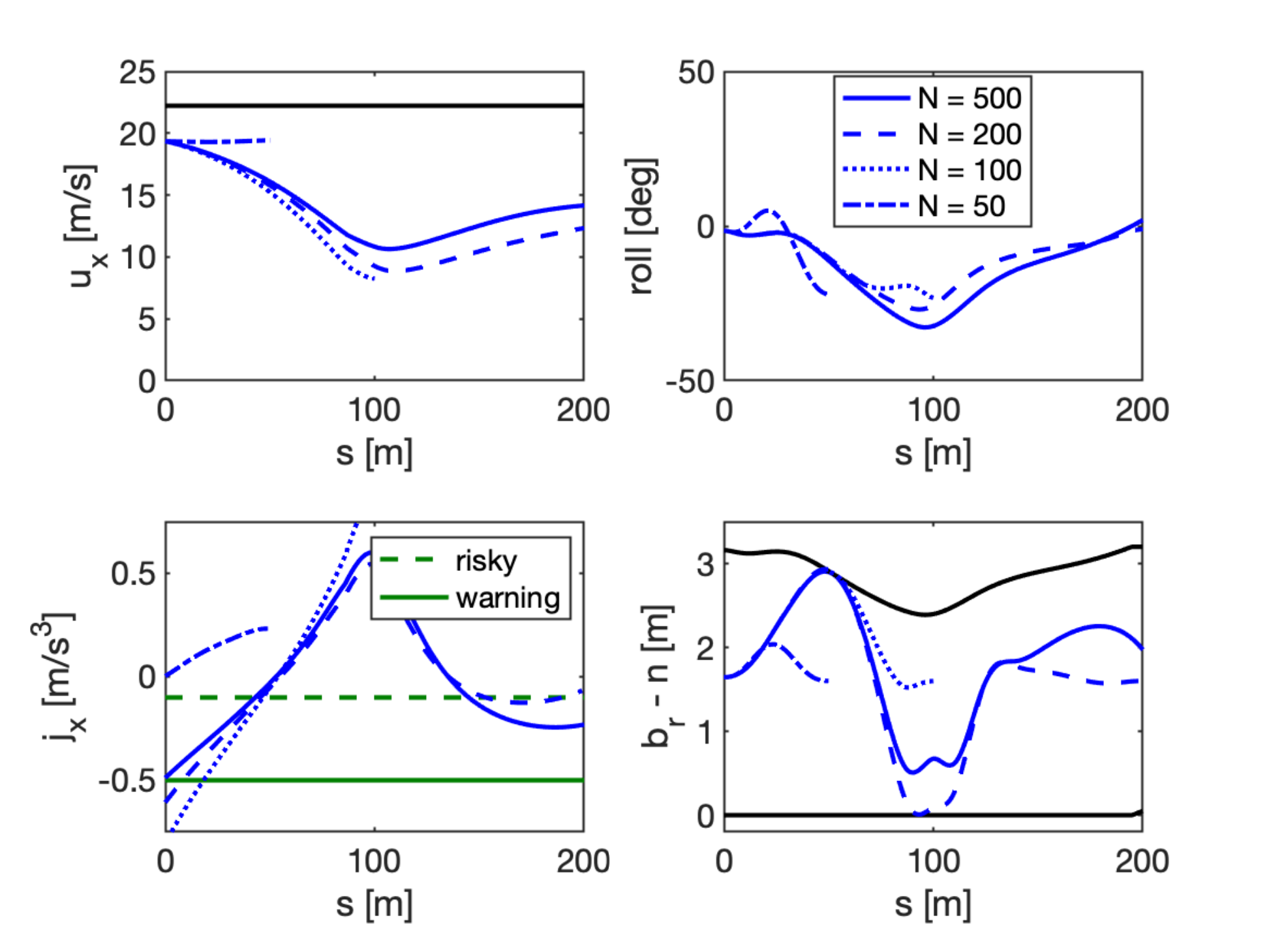}
	        \caption{Analyzing the influence of different horizon lengths on the optimal solution. }
	        \label{fig:trajPlot_horizon}
    	\end{minipage}
\end{figure*}

For the curve warning system evaluation we present a proof-of-concept case study. We highlight how the changes in the optimal control problem formulation have a significant and positive influence on the system performance. The case study evaluates the system performance on one road section of our \textit{Motorcycle Dataset}. The starting position is taken just before one of the most complex-to-maneuver curves of the dataset, precisely at the beginning of the braking maneuver, where a warning would be necessary. See Fig. \ref{fig:ablation_traj}, for the used curve section including a selection of the recorded on-board images and the prediction of \textit{RNet} and \textit{LNet}. In this scenario the motorcyclist is approaching a tight left S-curve, where the first curve is at the lowest point in terms of altitude. 

First, we investigate how the predicted behavior corresponds to the behavior of the rider, therefore we use a long prediction horizon of 500\,m, with a discretization distance of 1\,m. In Fig. \ref{fig:trajPlot} we can see the comparison between the rider and the optimal trajectory. On the top left we can see that the optimizer is planning a very similar velocity ($u_x$) through the first section but then approaches the speed limit more rapidly than the human rider. Likewise the roll angle is similar between the rider and the optimizer. When analyzing the (roll-corrected) intra-lane position, see bottom-left plot, we can observe that the optimal trajectory is using significantly more of the allowed road, but the general behavior is similar. We can also see that the rider is not hitting the apex at the correct spots, thus the system could also be used to teach the rider and improve their riding skills. Finally, we can see that we do not cause a warning, but are just at the boundary. 

In the bottom-right plot of Fig. \ref{fig:trajPlot}, where the intra-lane position is plotted, the effect of roll on the road constraint is shown, see Eq. \eqref{eq:roadCon}. This constraint prevents the riders head from crossing over the lane divider which is to be avoided. It is clearly visible that the effect cannot be neglected. The first reason being that the optimal control problem plans exactly to this limit, thus not including these constraints can render unsafe states as safe. Second, this constraint can render about a quarter of the road unsafe, which is a crucial safety consideration.

The second change we introduced is the effect of road slope on the longitudinal dynamics. To show the importance, we computed the optimal solution for the exact same location and the same setting with and without including the road slope, the comparison of the results can be seen in Fig. \ref{fig:trajPlot_slope}. Even thought the slope at this point is not steep, removing it still resulted in a significantly lower longitudinal jerk, thus, not including the slope will result in not warning the rider in potentially critical situations. 

The finial study we performed is investigating the influence of the prediction horizon. This is mainly important if the optimal control problem should be solved in real-time. As long as the horizon is at least 100\,m, which is the distance until the end of the first curve, the results look similar. With the general trend that shorter horizons lead to higher jerks, which is caused by the effect of the terminal constraint. However, when using a horizon of 50\,m the look ahead is too short and the optimal solution is suggesting to accelerate, this is a common problem with finite horizon approaches and can be solved with better terminal constraints \cite{Liniger2019, Novi2019}.

\section{CONCLUSION}

This paper has extended existing motorcycle curve warning systems by, for the first time, including intra-lane localization, adding a learned roll prediction approach, using industry standard maps for real-world evaluation and including more road attributes such as incline and speed limit in the controller formulation. The importance of these four objectives has been thoroughly discussed and mathematically formulated. This work made four contributions. We demonstrate that neural networks are the ideal candidates to solve the remaining problems of 1) intra-lane localization and 2) roll angle prediction in highly dynamic environments. To this end, we develop two CNNs able to solve both tasks respectively. Third, we reformulate the existing controller formulation to include additional road attributes leading to a more realistic model. Finally, previous work has largely relied on manually measuring or estimating road curvature and incline from images, we show that by including industry standard maps these safety systems can be scaled to various geographic regions. Extensive experiments have shown that our system is able to perform well \textit{in the wild}, giving rise to a more complete motorcycle curve warning system.

\section*{Acknowledgment}
\noindent
We are grateful for the support by HERE Technologies and Toyota Motor Europe.

\bibliographystyle{IEEEtran} 
\bibliography{egbib}

\end{document}